\documentclass[journal]{IEEEtran}

\usepackage{cite}
\usepackage{amsmath}
\usepackage{graphicx}
\usepackage{amsmath}
\usepackage{amssymb}
\usepackage{amsfonts}
\usepackage{cite}
\usepackage{booktabs}
\usepackage{amsmath}
\usepackage{multirow}

\usepackage{enumitem}
\usepackage{xcolor}
\usepackage{array}
\usepackage{bbding}

\usepackage{array}
\newcommand{\PreserveBackslash}[1]{\let\temp=\\#1\let\\=\temp}
\newcolumntype{C}[1]{>{\PreserveBackslash\centering}p{#1}}
\newcolumntype{R}[1]{>{\PreserveBackslash\raggedleft}p{#1}}
\newcolumntype{L}[1]{>{\PreserveBackslash\raggedright}p{#1}}

\begin{document}

    \title{Unified Open-Vocabulary Dense Visual Prediction}

    \author{Hengcan Shi,
    	Munawar Hayat,
    	and Jianfei Cai,   
        \thanks{H. Shi, M. Hayat, and J. Cai are with the Department of Data Science \& AI, Monash University, Melbourne 3800, Australia (e-mail: Hengcan.Shi@monash.edu; Munawar.Hayat@monash.edu; Jianfei.Cai@monash.edu).}
    }


    \maketitle

    \begin{abstract}
In recent years, open-vocabulary (OV) dense visual prediction (such as OV object detection, semantic, instance and panoptic segmentations) has attracted increasing research attention. However, most of existing approaches are task-specific and individually tackle each task. In this paper, we propose a Unified Open-Vocabulary Network (UOVN) to jointly address four common dense prediction tasks. Compared with separate models, a unified network is more desirable for diverse industrial applications. Moreover, OV dense prediction training data is relatively less. Separate networks can only leverage task-relevant training data, while a unified approach can integrate diverse training data to boost individual tasks. We address two major challenges in unified OV prediction. Firstly, unlike unified methods for fixed-set predictions, OV networks are usually trained with multi-modal data. Therefore, we propose a multi-modal, multi-scale and multi-task (MMM) decoding mechanism to better leverage multi-modal data. Secondly, because UOVN uses data from different tasks for training, there are significant domain and task gaps. We present a UOVN training mechanism to reduce such gaps. Experiments on four datasets demonstrate the effectiveness of our UOVN.
    \end{abstract}


\section{Introduction}\label{sec:intro}
Dense visual prediction tasks, including object detection, semantic, instance and panoptic segmentations, are important and fundamental computer vision problems.
They serve as crucial steps for many real-world applications, such as scene understanding \cite{ma2022learning, yang2022continual}, image generation \cite{zhao2023multi, liu2021intra} and multi-modal systems \cite{guo2021universal, shi2020query, zeng2021multi, shi2023unpaired}.
Traditional dense prediction methods \cite{DBLP:conf/cvpr/LiQDJW17, DBLP:conf/iccv/HeGDG17} are usually designed to recognize a fixed set of object categories. As a result, they have to be continually retrained to fit different real-world applications, because different applications normally involve varying category sets. Hence, open-vocabulary (OV) dense prediction \cite{du2022learning, minderer2022simple} has attracted increasing attention in recent years, where the model is trained to recognize an open set of object categories and thus can be directly used for diverse applications.

The existing OV approaches can be mainly divided into three categories.
The first is pre-training-based methods \cite{zareian2021open, gu2021open, ma2022open, xu2023open}, which learn dense predictions from traditional fixed-set dense data and leverage feature spaces from pre-trained vision-language models to recognize open-set objects. While these methods can generalize to open-vocabulary tasks, pre-trained models reduce the flexibility of these methods. They have to encode their features into the pre-trained feature space and cannot flexibly adjust them.
The second category is pseudo-label-based methods \cite{zhou2022detecting, gao2021towards, huynh2022open}, which first generate open-set dense pseudo labels from vision-language data, and then use these pseudo labels to train OV networks. Nevertheless, these methods suffer from the problem of inevitable noises in pseudo labels.
The third type \cite{kuo2022findit,li2022grounded} reformulates object detection and segmentation as referring grounding problems, and thus can leverage referring grounding training data to simultaneously enable the dense prediction and OV abilities. Such approaches avoid the noise in pseudo-label-based methods and are more flexible than pre-training-based methods. However, all these three types of methods only focus on one or several dense prediction tasks. 

The limitation of the existing OV methods motivates us to look for a unified model for all detection and segmentation tasks, which is more desirable for real-world applications. 
In addition, compared with traditional fixed-set dense prediction, the training data for OV dense prediction is relatively less. A unified network enables the integration of the training data across different tasks, and thus can enhance the individual performance of each task.
Note that there are several recent networks \cite{zhang2021k, wang2021max, cheng2021per, cheng2022masked, jain2023oneformer} successfully unifying traditional fixed-set detection and segmentation, demonstrating the feasibility of the unification. However, they cannot be directly used for OV scenarios, due to the lack of OV recognition and training mechanisms.

In particular, in this paper, we propose a Unified Open-Vocabulary Network (UOVN), including an MMM (multi-modal, multi-scale and multi-task) decoding mechanism to jointly detect and segment OV objects, and a UOVN training mechanism to reduce the domain and task gaps in OV training. We adopt the referring-grounding-based mechanism to recognize OV categories. 
During training, our inputs are an image and language queries which describe some objects in the image. Our MMM decoding generates bounding boxes and/or diverse segmentation masks. 
During inference, language queries can be replaced with object categories on the target dataset.
Inspired by traditional unification works \cite{zhang2021k, cheng2022masked}, our MMM decoding formulates all detection, semantic, instance and panoptic segmentations into a mask classification problem. We first extract instance-level and pixel-level embeddings, and then generate detection and segmentation results from them. 
More importantly, we present a multi-modal multi-scale deformable attention (MMDA) module to better extract embeddings based on vision-language information. 

\begin{figure*}
	\centerline{\includegraphics[scale=0.49]{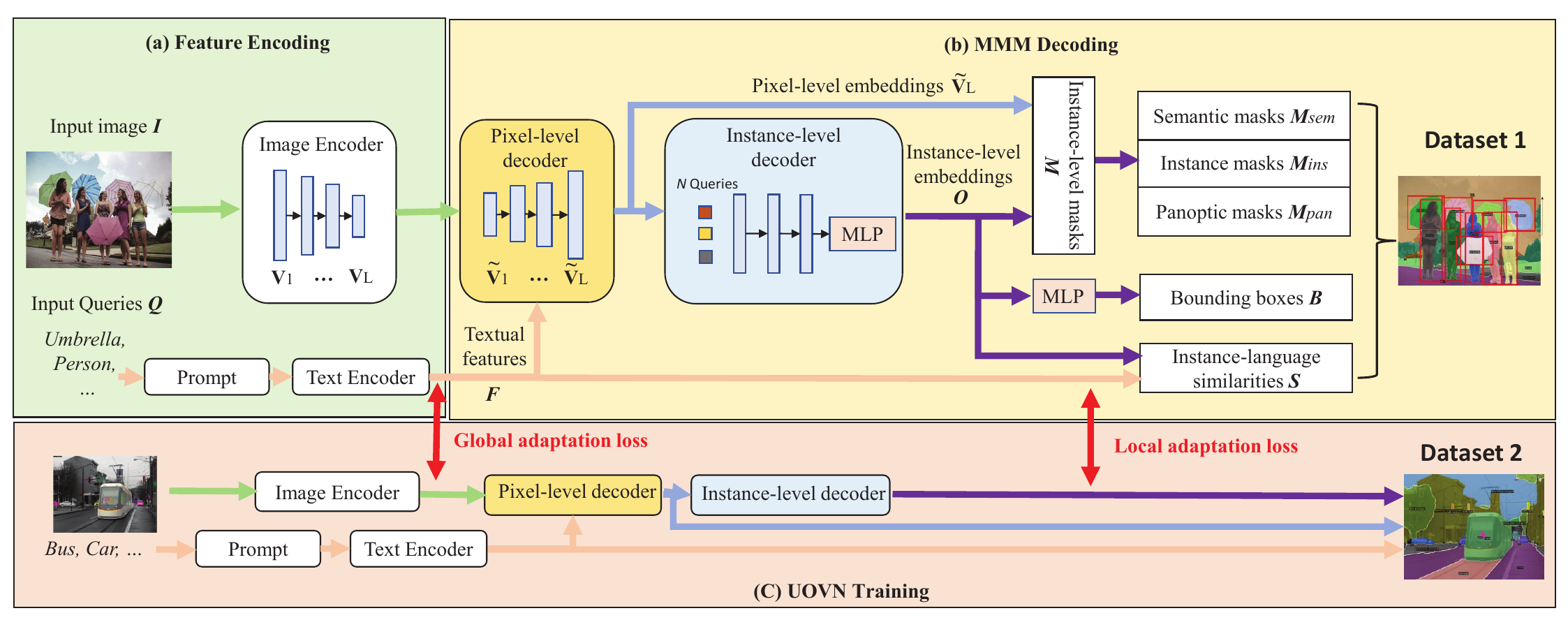}}
	\caption{Illustration of our UOVN. Our model consists of three parts. (a) Feature encoding, including an image encoder and a text encoder, extracts image and text features. (b) MMM (multi-modal, multi-scale and multi-task) decoding, containing a pixel-level decoder and an instance-level decoder, generates OV detection and segmentation results. We also propose a multi-modal multi-scale deformable attention (MMDA) module in this part, to better model multi-modal information. (c) UVON training leverages different datasets to train our network. We introduce global and local adaptation losses to extract domain-invariant visual embeddings.}
	\label{fig_method}
\end{figure*}

We use referring detection and segmentation training data for our model. There are significant domain and task gaps. Therefore, we propose a UOVN training mechanism to address such gaps.  We expect to learn domain-invariant embeddings for training samples from different datasets. Traditional domain adaptation methods \cite{xu2021cdtrans, xie2019multi} usually input training samples in the same category into a batch, and make their features similar. However, in our training data, we only have language queries available instead of categories. 
We observe that if two images have similar  queries, they  likely contain similar content.
Based on such observation, we introduce global and local adaptation losses, which essentially leverage the textual similarity to guide the visual learning to generate domain-invariant embeddings.

In summary, our major contributions are as follows. 
\begin{enumerate}
	\item We propose a novel unified network for OV object detection and diverse segmentation tasks.
	\item We introduce an MMM decoding and a UOVN training mechanism. 
	The MMM decoding including an MMDA module leverages multi-modal information and the mask classification architecture to enable OV detection, semantic, instance and panoptic segmentations. The UOVN training mechanism allows global and local adaptions among various datasets.
	\item Extensive experiments on COCO, LVIS, ADE20k and Pascal Context show that our network achieves significant improvements.
\end{enumerate}

\section{Related Work}
\subsection{Unified Networks for Detection and Segmentation} 

Object detection, semantic, instance and panoptic segmentations are typical dense visual prediction tasks and widely used in practice. Detection methods \cite{ren2015faster, DBLP:conf/cvpr/RedmonDGF16, liu2016ssd} localize and classify objects in an image. Semantic segmentation methods \cite{long2015fully, shi2019scene, chen2018encoder, shi2023transformer} determine the mask for each object category, which is achieved usually by classifying each pixel in the image. Unlike semantic segmentation, instance segmentation approaches \cite{DBLP:conf/cvpr/LiQDJW17, DBLP:conf/iccv/HeGDG17} predict masks for different object instances. Panoptic segmentation methods generate instance-level masks for foreground `thing' objects and category-level masks for background `stuff' objects. Earlier panoptic segmentation approaches \cite{kirillov2019panoptic1, kirillov2019panoptic2} use two networks to separately predict instance-level and category-level masks. UPSNet \cite{xiong2019upsnet} designs a shared encoder with different segmentation heads to solve the panoptic segmentation problem. Since both detection and instance segmentation are instance-level prediction problems, many works \cite{DBLP:conf/cvpr/DaiHS16, DBLP:conf/iccv/HeGDG17} use a single instance-level network to unify them.
To unify segmentation tasks, recent works \cite{zhang2021k, wang2021max, cheng2021per, cheng2022masked, jain2023oneformer} reformulate different segmentation tasks as a mask prediction problem. Several methods \cite{chen2022pix2seq, chen2022unified, kolesnikovuvim} consider detection and segmentation as a (text) sequence prediction problem to flexibly unify them with other tasks, but they do not outperform mask-prediction-based approaches.
These fixed-category unified approaches are foundations of our work. However, they cannot be directly used in OV tasks, due to the lack of OV recognition and training mechanisms.

\subsection{Open-Vocabulary Detection and Segmentation} 
The existing OV detection and segmentation works can be generally categorized into three types. The first is pre-training-based approaches. They normally train their models with fixed-set detection/segmentation data to segment/localize objects, and simultaneously distill knowledge from pre-trained vision-language models to recognize OV categories.
For example, \cite{zareian2021open, dong2022maskclip, xu2023learning} use image-caption data to pre-train models for OV detection \cite{zareian2021open} and semantic segmentation \cite{dong2022maskclip, xu2023learning}.
Many methods \cite{gu2021open, ma2022open} employ off-the-shelf vision-language pre-trained models, such as CLIP \cite{radford2021learning}. 
ViLD \cite{gu2021open} and OV-DETR \cite{zang2022open} distill knowledge from CLIP to Mask RCNN \cite{DBLP:conf/iccv/HeGDG17} and DETR \cite{carion2020end}, respectively, to predict open-vocabulary detection results. HierKD \cite{ma2022open} further embeds multi-level object features and global image features into the CLIP feature space. F-VLM \cite{kuo2022f} add simple classification heads and localization heads to CLIP, and fine-tune their models on detection data. DetPro \cite{du2022learning} incorporates prompt learning to boost the performance, and Promptdet \cite{feng2022promptdet} further enhances the prompt learning to region level. Promptdet \cite{feng2022promptdet} is also trained with web images to obtain more knowledge. 
MaskCLIP-panoptic \cite{ding2022open} designs mask class tokens and relative mask attention to better distill CLIP knowledge for OV panoptic segmentation. ODISE \cite{xu2023open} employs pre-trained diffusion models to learn open-vocabulary knowledge.
These methods successfully leverage vision-language pre-training to improve their OV recognition ability. Nevertheless, their flexibility is limited by pre-trained models in the sense that they have to encode their features into the pre-trained feature space and cannot flexibly adjust them.

The second type is pseudo-label-based methods that generate pseudo detection/segmentation labels from image-level OV data, to address the issue of lacking dense OV annotations.
In particular, Detic \cite{zhou2022detecting} generates pseudo bounding boxes from large image classification datasets.RegionCLIP \cite{zhong2022regionclip} and VL-PLM \cite{zhao2022exploiting} extract pseudo boxes from CLIP by using class activation maps (CAMs) or pre-trained RPN \cite{ren2015faster}. VLDet \cite{lin2022learning} predicts nouns from image captions and aligns each noun to object proposals generated by pre-trained object detectors.
MaskCLIP+ \cite{zhou2021denseclip} uses CLIP visual encoder to extract pseudo masks for OV semantic segmentation. XPM \cite{huynh2022open} generates pseudo masks from image captions for instance segmentation. Instead of directly using pseudo labels, OpenSeg \cite{ghiasi2021open} and Rasheed \emph{et al.} \cite{rasheed2022bridging} first output dense predictions and then restore image-level results from these predictions to train their models. Nevertheless, pseudo-labels introduce inevitable noise during learning.

The third category of approaches, referring-grounding-based works, point out the high similarity between OV dense prediction and grounding, and use grounding frameworks to tackle OV dense prediction. Since grounding data includes detection/segmentation annotations for diverse objects, FindIt \cite{kuo2022findit} combines grounding and object detection data to train a model, which shows OV detection ability. X-DETR \cite{cai2022x} reformulates object detection and grounding as an instance-text alignment problem, and designs a unified alignment network for both tasks. GLIP \cite{li2022grounded} enhances the vision-language interaction in the alignment framework, and also collects millions of data for training. GLIPv2 \cite{zhang2022glipv2} extends GLIP for more tasks, such as instance segmentation and image captioning. X-Decoder \cite{zou2022generalized} integrates grounding, image caption and image segmentation data for OV segmentation. However, all these three types of works cannot simultaneously deal with detection, semantic, instance and panoptic segmentation tasks. Unlike them, we propose a unified network incorporating the developed MMM decoding and UOVN training mechanisms for all these tasks.

\section{Proposed Method}\label{analysis}

\subsection{Problem Definition and Method Overview}
We formulate OV detection/segmentation as a proposal/mask-language alignment problem. Specifically, our inputs are an image $\mathbf{I} \in \mathbb{R}^{H \times W \times D_{I}}$ and language queries $\mathbf{Q} \in \mathbb{R}^{C \times P \times D_{Q}}$, where $H$ and $W$ are the height and width of the image, $D_{I}$ and $D_{Q}$ are respectively the channel dimensions of the image and text queries, and $C$ is the number of language queries for the image. During training with referring grounding/segmentation data,
$C$ is the number of referring descriptions for an image, where each description can be a word, a phrase or a sentence, describing an object in the image. During inference, $C$ is the number of object categories on the target dataset. $P$ represents the number of words in each language query.

Our method contains three modules, as illustrated in Fig.~\ref{fig_method}. (a) \emph{Feature encoding}: an image encoder and a text encoder firstly extract image and text features. (b) \emph{MMM (multi-modal, multi-scale and multi-task) decoding}: in this module, we first output $N$ instance-level bounding boxes, masks and corresponding visual embeddings. Then, $C$ textual embeddings are generated for language queries. After that, we construct a  vision-language similarity matrix $\mathbf{S} \in \mathbb{R}^{N \times C}$, which measures the similarity between each object visual embedding and each query textual embedding. The category for each object can be determined by this matrix. Finally, diverse dense predictions are output based on the $N$ instance-level boxes, masks and classification results. (c) \emph{UOVN training}: since various datasets are used during training, we propose global and local adaptation losses to reduce the domain and task gaps among these datasets.
Next, we introduce each module in detail.

\subsection{Feature Encoding} \label{Ecoding}
\textbf{Language query encoder.} For input queries $\mathbf{Q} \in \mathbb{R}^{C \times P \times D_{Q}}$, we leverage a textual encoder (e.g., BERT \cite{kenton2019bert} or RoBERTa \cite{liu2019roberta}) to generate a textual feature map $\mathbf{F} \in \mathbb{R}^{C \times D_{F}}$. In this feature map, each vector $\mathbf{f}_{c} \in \mathbb{R}^{D_{F}} (c=1,...,C)$ encodes a language query, and $D_{F}$ is its dimension. Prompt technologies, such as the hand-crafted prompt `a photo of a [Query]' \cite{radford2021learning} or learnable prompt \cite{zhou2022learning}, can be used to further enhance textual features.

\textbf{Image encoder.} We use an image encoder (ResNet \cite{he2016deep} or Swin Transformer \cite{liu2021swin}) to extract multi-scale feature maps $ \{ \mathbf{V}_{l} \in \mathbb{R}^{H_{l} \times W_{l} \times D_{l}} \}_{l=1}^{L} $ for the input image, where $L$ is the number of scales. Feature maps of different scales are extracted from different stages of the image encoder. $H_{l}$, $W_{l}$ and $D_{l}$ are the height, width and channel number for the $l$-th feature map, respectively. 

\begin{figure*}
	\centerline{\includegraphics[scale=0.2]{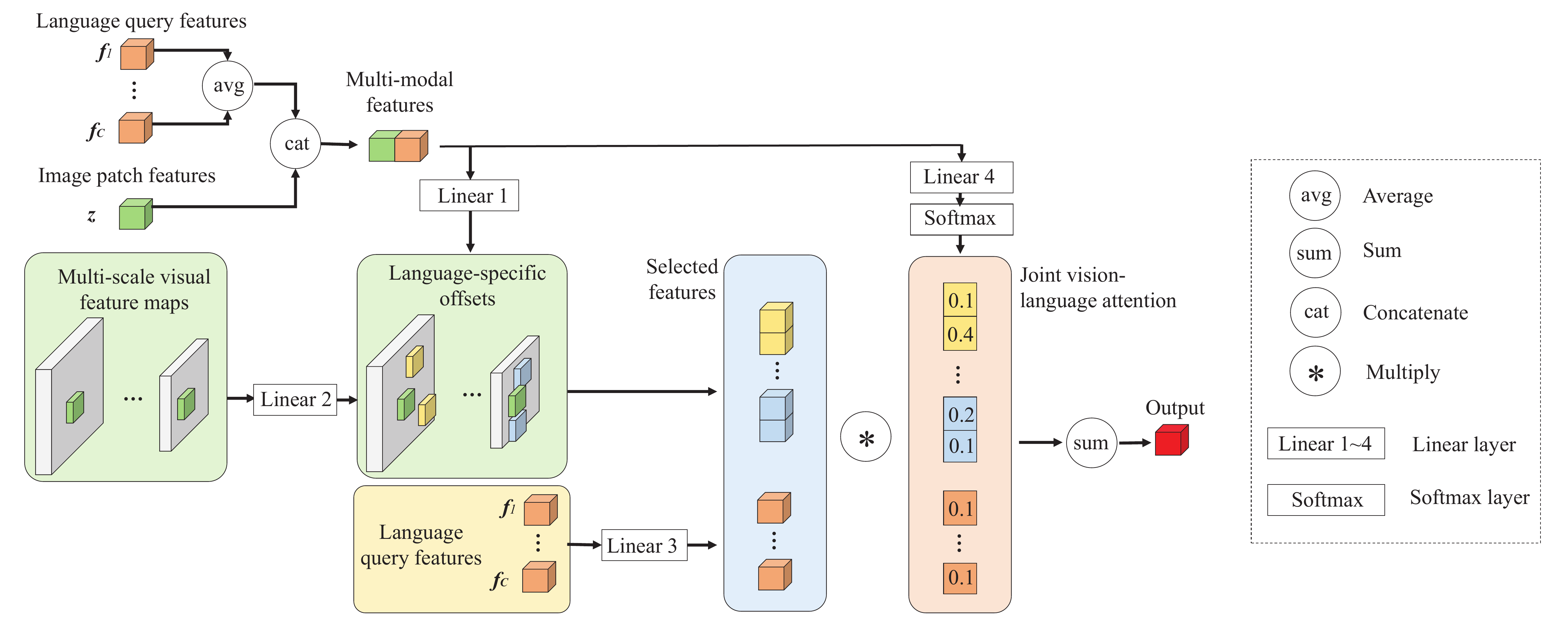}}
	\caption{Illustration of our multi-modal multi-scale deformable attention (MMDA) module, with one attention head for simplicity. We integrate both vision and language cues to predict offsets in deformable attention \cite{zhu2021deformable}, and thus our MMDA can model language-specific visual relationships. Meanwhile, we learn joint vision-language attentions to fuse multi-modal features for better decoding.}
	\label{fig_method2}
\end{figure*}

\subsection{Multi-modal Multi-scale Multi-task Decoding} \label{Decoding}
We expect to design a unified and clean decoding mechanism for OV dense prediction tasks. Inspired by the recent unified framework Mask2Former \cite{cheng2022masked}, we follow its two-decoder architecture: a pixel-level decoder and an instance-level decoder to generate pixel- and instance-level embeddings, respectively. The results of different tasks can be formulated as different combinations of these embeddings.

\textbf{MMDA module.} In Mask2Former \cite{cheng2022masked}, the pixel-level decoder is a vision transformer to integrate multi-scale image features and generate pixel-level embeddings. However, in OV scenarios, language cues are also important. Thus, we propose a multi-modal multi-scale deformable attention (MMDA) module, as shown in Fig.~\ref{fig_method2}, to generate multi-scale pixel-level embeddings based on both vision and language information. In particular, considering there may be multiple language query features, (e.g., 80 queries on the COCO dataset during inference), we first aggregate all language query features $\{ \mathbf{f}_{c} \}_{c=1}^{C}$ into a single feature vector by average pooling. Then, the aggregated query feature is concatenated with each image patch feature. The concatenated multi-modal feature is used to predict offsets and compute attention maps in multi-scale deformable attention \cite{zhu2021deformable}. 

Unlike original self-attention, which models relationships among all image patches, deformable attention \cite{zhu2021deformable} only models the relationships among $K$ selected patches for every image patch $\mathbf{z}$ to reduce computational costs.
It uses a linear layer (`Linear 1' in Fig.~\ref{fig_method2}) to predict $K$ offsets to select patches.
In our MMDA, the $K$ offsets are predicted by both vision and language cues to model \emph{language-specific visual relationships}. For multi-scale image features, we select $K$ patches on each scale. Finally, for every image patch $\mathbf{z}$, a set of $K \times L$ selected patch feature vectors is constructed. To further model vision-language information, we also add language query features to this set. Note that all features in this set are transformed by linear layers ('Linear 2\&3' in Fig.~\ref{fig_method2}) as `values' in attention model.

A linear layer ('Linear 4' in Fig.~\ref{fig_method2}) and a softmax are used to calculate attention maps. These attention maps measure the importance of every feature vector in the selected feature set, where we also leverage both vision and language cues to learn \emph{joint vision-language attention}. The output of one attention head is a weighted sum of features in the set, where the weights are from the attention map. In each Transformer layer, we use eight attention heads and integrate their outputs into one feature vector.

\textbf{Detection and segmentation output generation.} The outputs of the pixel-level decoder are transformed feature maps $\{\mathbf{\tilde{V}}_{l} \in \mathbb{R}^{H_{l} \times W_{l} \times D_{F}} \}_{l=1}^{L} $, where $D_{F}$ is their dimension,  same as the dimension of textual features. Each of these feature maps incorporates multi-scale vision and language information. Similar to Mask2Former \cite{cheng2022masked}, we use a Transformer decoder with masked attention and a two-layer MLP (Multilayer perceptron) as our instance-level decoder. It takes multi-scale features maps $\{\mathbf{\tilde{V}}_{l} \}_{l=1}^{L-1}$
generated from the pixel-level decoder as key and value, and includes $N$ learnable Transformer queries. The outputs are instance-level embeddings $\mathbf{O} \in \mathbb{R}^{N \times D_{F}}$, where $D_{F}$ is the dimension. Each feature vector $\mathbf{o}_{n} \in \mathbb{R}^{D_{F}} (n=1,...,N)$ in $\mathbf{O}$ embeds an object instance.

We then generate dense predictions from the pixel- and instance-level embeddings $\mathbf{\tilde{V}}_{L}$ and $\mathbf{O}$. Specifically, segmentation masks can be predicted as
\begin{equation}
	\mathbf{M} = Binary(Sigmoid(\mathbf{O}\mathbf{\tilde{V}}_{L}^{T}))
\end{equation} 
where $T$ means a tensor transposition, and $Sigmoid(\cdot)$ and $Binary(\cdot)$ are sigmoid and binary thresholding functions, respectively. In the binary thresholding function, if a value is higher than 0.5, we set it to 1; otherwise 0. 
$\mathbf{M} \in \mathbb{R}^{N \times H_{L} \times W_{L} } $ are segmentation masks for $N$ object instances and each mask is a binary one of $H_{L} \times W_{L}$. These segmentation masks can be directly used as \textit{instance segmentation} results $\mathbf{M}_{ins}$.

The similarity matrix $\mathbf{S} \in \mathbb{R}^{N \times C} $ between visual instances and language queries can be calculated as
\begin{equation} \label{eq:s}
	\mathbf{S} = \mathbf{O}\mathbf{F}^{T}.
\end{equation} 
During inference, language queries are object categories of the target dataset. We can also add a `no object' class similar to fixed-set works \cite{carion2020end, cheng2021per}. For each object instance, we find the class with the highest similarity as its classification result. However, different from fixed-set datasets, in referring grounding/segmentation datasets, an object instance may correspond to multiple language queries, and a language query may describe multiple object instances. Therefore, during training, we calculate binary cross-entropy loss for every element in $\mathbf{S}$.

\textit{Semantic segmentation} masks $\mathbf{M}_{sem} \in \mathbb{R}^{C \times H_{L} \times W_{L}} $ can be treated as a combination of instance-level segmentation masks $\mathbf{M}$. Through Eq.~(1)\&(2), we obtain instance-level masks and their classes. For every class, we fuse instance-level masks of this class as the semantic segmentation results. \textit{Panoptic segmentation} expects to generate instance segmentation results for pre-defined classes, while semantic segmentation results for other classes. Therefore, we can combine instance and semantic segmentation results based on their classes as panoptic segmentation results $\mathbf{M}_{pan}$. 

A simple way to predict \textit{object detection} results is via a post-processing step to directly generate bounding boxes from instance segmentation masks. However, to better leverage bounding-box-only annotations during training, we add an additional two-layer MLP to generate bounding boxes $\mathbf{B} \in \mathbb{R}^{N \times 4}$ based on instance-level embeddings $\mathbf{O}$.

\subsection{UOVN Training} \label{Training}
We train our network jointly with segmentation, detection, classification and adaptation losses:
\begin{equation}
	L  = \lambda_{1}L_{seg} + \lambda_{2}L_{det} + \lambda_{3}L_{cls} + \lambda_{4}L_{adp},
\end{equation}
where $L_{seg}$ is for mask predictions, $L_{det}$ is the detection loss for the bounding box regression, $L_{cls}$ is for OV classification,
and $L_{adp}$ is for the adaptation in cross-dataset training.  $\lambda_{1}, \lambda_{2}, \lambda_{3}$ and $\lambda_{4}$ are hyperparameters to weight different losses. Note that our losses can be flexibly adjusted according to the training data, e.g., the segmentation loss can be removed when only box annotations are given.

Specifically, we use the common detection loss \cite{carion2020end} as our $L_{det}$, which minimizes the \emph{smooth $l_1$ distance} and \emph{GIoU} between ground truths and  bounding box predictions $\mathbf{B}$. The segmentation loss in \cite{cheng2021per, cheng2022masked} is employed as our $L_{seg}$ to optimize the \emph{binary cross-entropy} and \emph{Dice} between  mask predictions $\mathbf{M}$ and ground truths. Bipartite Hungarian matching is used to align predicted boxes/masks with ground truths. The OV classification loss $L_{cls}$ is a binary cross-entropy between the similarity matrix and the ground truth matrix:
\begin{equation}
	L_{cls} = L_{BCE}(\mathbf{S}_{gt}, Sigmoid(\mathbf{S}))
\end{equation}
where the ground truth (GT) matrix $\mathbf{S}_{gt} \in \mathbb{R}^{N \times C}$ is generated from mask/box labels and their corresponding language queries. Specifically, we set GT similarity $s_{n,c}$ to 1, when the IoU between the $n$-th predicted mask/box and the GT mask/box of the $c$-th query is larger than 0.5; otherwise, we set it to 0.

\textbf{Adaption loss.} The adaption loss $L_{adp}$ is proposed to learn domain-invariant embeddings to reduce cross-dataset gaps, which consists of two parts: $L_{adp}  = L_{adp-g} +  L_{adp-l}$.  $L_{adp-g}$ is a global adaption loss to capture image-level domain-invariant embeddings, while $L_{adp-l}$ represents a local adaption loss for learning instance-level domain-invariant embedding. Fixed-set domain adaptation methods \cite{xu2021cdtrans, xie2019multi} often extract domain-invariant features by inputting images from different datasets but in the same category into a batch, and making their features similar. However, our training data only includes language queries rather than image categories. We notice that if two images have similar queries, they likely share similar content,
and vice versa. Based on this observation, we propose to use the textual similarity to guide the visual similarity learning for domain adaptation.

Concretely, we first input two image-query pairs from different datasets into a batch. Then, we compute the visual similarity $a_v$ between the images and the textual similarity $a_l$ between language queries. Considering an image may have multiple language queries, we average query features $\{ \mathbf{f}_{c} \}_{c=1}^{C}$ into a single vector $\mathbf{f} \in \mathbb{R}^{D_{F}}$. The visual feature map $\mathbf{V}_{L} \in \mathbb{R}^{H_{L} \times W_{L} \times D_{L}}$ in the last stage of the image encoder is also averaged into a vector $\mathbf{v} \in \mathbb{R}^{D_{L}}$. 
After that, we compute $a_v$ as the cosine similarity between the averaged visual features of the two images, and $a_l$ as the cosine similarity between the two averaged query features. The global adaption loss $L_{adp-g}$ then matches the visual and textual similarities:
\begin{equation}\label{eq:adp-g}
	L_{adp-g}  = L_{dis}(a_v, a_l).
\end{equation}
Since $a_v$ and $a_l$ are two numbers, we use the $\ell_1$-norm distance as our loss function $L_{dis}$. We encourage the visual embeddings of two images from different datasets to have a similar distance as that of their queries embeddings.

Similarly, we can apply the same adaptation idea at the instance level. For two objects from different datasets, we make their visual similarity to be close to the textual similarity between their corresponding queries.
In particular, we first generate the language similarity matrix $\mathbf{A_L} \in \mathbb{R}^{C_{1} \times C_{2}}$ between the two training samples from different datasets: $\mathbf{A_L} = \mathbf{F}^{1}(\mathbf{F}^{2})^{T}$, where
$\mathbf{F}^{1} \in \mathbb{R}^{C_{1} \times D_{F}}$ and $\mathbf{F}^{2} \in \mathbb{R}^{C_{2} \times D_{F}}$ are the language query feature maps in the two training samples, respectively, and $C_{1}$ and $C_{2}$ are their numbers of queries. Then, based on the cross-modal similarity matrix $\mathbf{S}$ in Eq.\ref{eq:s}, we can get the instance embedding corresponding to each language query. If a query corresponds to multiple instances, we average their embeddings. In this way, we can obtain two instance embedding matrices $\mathbf{\tilde{O}}^{1} \in \mathbb{R}^{C_{1} \times D_{F}}$ and $\mathbf{\tilde{O}}^{2} \in \mathbb{R}^{C_{2} \times D_{F}}$ for the two training samples, respectively, and compute the visual similarity matrix $\mathbf{A_V}= \mathbf{\tilde{O}}^{1}(\mathbf{\tilde{O}}^{2})^{T}$, where $\mathbf{A_V} \in \mathbb{R}^{C_{1} \times C_{2}}$. The local adaptation loss $L_{adp-l}$ is then given by
\begin{equation}
	L_{adp-l}  = L_{KL}(\mathbf{A_L}, \mathbf{A_V})
\end{equation}
where we use a Kullback-Leibler divergence loss $L_{KL}$ to constrain the similarity between the two matrices $\mathbf{A_L}$ and $\mathbf{A_V}$.

\begin{table*}
	\centering
	\caption{Cross-dataset results on COCO validation. `Det', `Ins' and `Pan' are object detection, instance and panoptic segmentations, respectively. `\scalebox{0.85}[1]{$\times$}' means the method cannot deal with this task. `SD' is stable diffusion model. `Model A' is a modified Mask2Former \cite{cheng2022masked}, where we replace the classification part with a vision-language similarity matrix for OV recognition. Our UOVN is able to deal with all detection and segmentation tasks, and outperforms most of the previous approaches. GLIP \cite{li2022grounded} use larger backbones and much more training/pre-training data. X-Decoder \cite{zou2022generalized} and ODISE \cite{xu2023open} are fully-supervised on COCO.}
	\scalebox{1.0}{
		\begin{tabular}{l|c|c|c|c|c|c}
			\toprule
			& &\multicolumn{2}{c|}{Supervision} & Det  & Ins & Pan\\
			\hline
			Method & Backbone & VL Pre-training & Training & $mAP$ & $mAP$ & $PQ$ \\
			\hline
			\midrule
			ViLD \cite{gu2021open}     & Res-50  & 400M (CLIP) & 100K (LVIS base) & 36.6 & \scalebox{0.85}[1]{$\times$} & \scalebox{0.85}[1]{$\times$} \\
			DetPro \cite{du2022learning} & Res-50 & 400M (CLIP) & 100K (LVIS base) & 34.9 & \scalebox{0.85}[1]{$\times$} & \scalebox{0.85}[1]{$\times$} \\
			OV-DETR \cite{zang2022open} & Res-50 & 400M (CLIP) & 100K (LVIS base) & 38.1 & \scalebox{0.85}[1]{$\times$} & \scalebox{0.85}[1]{$\times$} \\
			X-DETR \cite{cai2022x}      & Res-101 & -  & 14M (grounding, detection, caption) & 26.5 & \scalebox{0.85}[1]{$\times$} & \scalebox{0.85}[1]{$\times$} \\
			GLIP \cite{li2022grounded}    & Swin-L & - & 27M (grounding, detection, caption) & \textbf{49.8} & \scalebox{0.85}[1]{$\times$} & \scalebox{0.85}[1]{$\times$} \\
			\textcolor{gray}{X-Decoder \cite{zou2022generalized} (fully-supervised)} & \textcolor{gray}{Focal-T} & \textcolor{gray}{-} & \textcolor{gray}{4M (caption, grounding, COCO Pan)} & \textcolor{gray}{\scalebox{0.85}[1]{$\times$}} & \textcolor{gray}{46.7} & \textcolor{gray}{56.9} \\
			\textcolor{gray}{ODISE \cite{xu2023open} (fully-supervised)} & \textcolor{gray}{SD} & \textcolor{gray}{400M (SD) + 400M (CLIP)} & \textcolor{gray}{240K (COCO Cap, COCO Pan)} & \textcolor{gray}{\scalebox{0.85}[1]{$\times$}} & \textcolor{gray}{\scalebox{0.85}[1]{$\times$}} & \textcolor{gray}{55.4} \\
			\hline
			GLIP \cite{li2022grounded} (re-train) & Res-50 & - & 300K (grounding, LVIS base) & 35.4 & \scalebox{0.85}[1]{$\times$} & \scalebox{0.85}[1]{$\times$} \\
			Model A (modified Mask2Former \cite{cheng2022masked}) & Res-50 & - & 300K (grounding, LVIS base) & \scalebox{0.85}[1]{$\times$} & 20.8 & 24.4 \\
			UOVN (Ours) & Res-50 & - & 300K (grounding, LVIS base)                & 41.3  & \textbf{30.5} & \textbf{32.7} \\				
			\bottomrule	
	\end{tabular}}
	\label{tab_result_coco}
\end{table*}

\section{Experiments}

\subsection{Experiment Settings}
Following previous OV methods \cite{gu2021open, ding2022open, du2022learning}, we evaluate UOVN at two scenarios: \emph{cross-dataset} and \emph{zero-shot}. Note that there is no single training setting used across existing OV methods, since OV methods focus on using unrestricted training data to recognize unseen objects. Most of the existing approaches collect different training data and/or pre-training models. For a fair comparison, we reproduce key baselines on our training data.

\textbf{Cross-dataset scenario.} This scenario uses different training and testing datasets to gauge the OV ability. To train our model, we combine widely used referring grounding/segmentation datasets: RefCOCO \cite{yeh2018interpretable}, ReferItGame \cite{Kazemzadeh2014ReferItGame}, Visual Genome \cite{krishna2016visual} and PhraseCut \cite{wu2020phrasecut}. As prior works \cite{gu2021open, du2022learning, zang2022open} usually use LVIS base (886 common and frequent classes) \cite{gupta2019lvis} for training, we also include these data in our training set. We have 300K training samples in total.  
Our goal is to verify the effectiveness of our unified network rather than train a large pre-training model. Thus, we do not use millions of training data like pre-training works \cite{li2022grounded}. Following prior works \cite{gu2021open, ding2022open}, we take COCO \cite{lin2014microsoft}, ADE20K \cite{zhou2017scene} and Pascal Context \cite{mottaghi2014role} as testing datasets. There are overlapped images between COCO and LVIS/grounding datasets. Although most of the prior works use these overlapped images during training, we remove them from our training data to better verify the OV ability. 

\textbf{Zero-shot scenario.} In this scenario, the training and testing are on the same dataset, but with disjoint classes. We report zero-shot results on COCO \cite{lin2014microsoft} and LVIS \cite{gupta2019lvis}. For the COCO \cite{lin2014microsoft} dataset, we use the grounding data and COCO base (48 base classes) for training, while testing our model on 17 COCO novel classes. For LVIS \cite{gupta2019lvis}, our training data the combination of grounding data and LVIS base (886 common and frequent classes). The testing set is 337 LVIS rare classes. We not only exclude overlapped images from our training data, but also remove language queries which contain target classes in testing sets.

\textbf{Metrics.} We use $mIoU$ for semantic segmentation, $mAP$ and $AP^{50}$ for detection and instance segmentation, as well as $PQ$ for panoptic segmention \cite{kirillov2019panoptic1}.

\subsection{Dataset Details }
\textbf{Grounding datasets.} We use RefCOCO \cite{yeh2018interpretable}, ReferItGame \cite{Kazemzadeh2014ReferItGame}, Visual Genome \cite{krishna2016visual} and PhraseCut \cite{wu2020phrasecut} grounding data to enable the OV recognition ability. Both RefCOCO \cite{yeh2018interpretable} and ReferItGame \cite{Kazemzadeh2014ReferItGame} contain 20K images. Visual Genome \cite{krishna2016visual} and PhraseCut \cite{wu2020phrasecut} have 108K and 77K images, respectively. Visual Genome \cite{krishna2016visual} provides bounding box annotations with phrase and word descriptions. RefCOCO \cite{yeh2018interpretable} contains both bounding box and segmentation annotations, but only `things' objects are labeled. ReferItGame \cite{Kazemzadeh2014ReferItGame} and PhraseCut \cite{wu2020phrasecut} provide bounding box and segmentation annotations for both `things' and `stuff' objects. 

\textbf{COCO.} COCO 2017 \cite{lin2014microsoft} contains 120K training and 5K validation images for the detection, instance and panoptic segmentation tasks. There are 80 object classes for detection and instance segmentation, as well as 133 object classes in panoptic segmentation. 48 base classes and 17 novel classes are selected as the same in \cite{bansal2018zero} for zero-shot evaluation.
 
\textbf{ADE20K.} We use the ADE20K \cite{zhou2017scene} dataset during testing. 2K validation images in this dataset are annotated for semantic and panoptic segmentaions, and objects are divided into 847 or 150 classes. 

\textbf{Pascal Context.} The Pascal Context \cite{mottaghi2014role} dataset is a semantic segmentation benchmark. There are 5.1K validation images and two kinds of class definitions: 459 and 59 classes.

\textbf{LVIS.} LVIS \cite{gupta2019lvis} includes object detection and instance segmentation annotations for 100K training and 20K validation images. Object categories in LVIS \cite{gupta2019lvis} are split into three sets: 405 frequent, 461 common and 337 rare classes. We use 886 frequent and common classes as base classes for training, while taking 337 rare classes for zero-shot testing.

\subsection{Implementation Details}
We choose RoBERTa \cite{liu2019roberta} as our text encoder and add a linear layer to convert the textual feature dimension $D_{F}$ to $256$. We fix RoBERTa during training and only update the parameters of the linear layer. We do not use any prompt during training, only the prompt `A photo of a [query]' is used during inference.
Res-50 \cite{he2016deep} pre-trained on ImageNet is used as our image encoder, because most of the previous methods \cite{gu2021open, huynh2022open, ding2022open} employ it. Feature maps on four scales (i.e., $L=4$) are extracted. In our MMDA module, we set the number $K$ of patches to 4. In the instance-level decoder, 100 instance embeddings are generated, i.e., $N=100$. For the adaptation training, we randomly input two images from different datasets into a batch. $\lambda_{1}$, $\lambda_{2}$, $\lambda_{3}$ and $\lambda_{4}$ are simply set to 2.0, 2.0, 1.0 and 1.0, respectively, and fixed for all datasets. 
Other network and training settings are the same as Mask2former \cite{cheng2022masked}.
Similar to \cite{li2022grounded}, the maximum number of language queries is set to 256. If an image corresponds to over 256 language queries, we divide them into multiple sets and separately generate results for each set. All experiments are conducted on the Pytorch platform \cite{paszke2019pytorch} with 8 V100 GPUs and 2 images per GPU.

\begin{table*}
	\centering
	\caption{Cross-dataset results on ADE20K and Pascal Context validation. `A-847' and `A-150' mean 847 and 150 classes on ADE20K, respectively, while `P-459' and `P-59' mean 459 and 59 classes for Pascal Context. `Sem' and `Pan' are semantic and panoptic segmentation. `\scalebox{0.85}[1]{$\times$}' means the method cannot deal with this task. `SD' is stable diffusion. Only ADE20K contains panoptic segmentation validation.}	
	\scalebox{0.85}{
		\begin{tabular}{l|c|c|c|cccc|c}
			\toprule
			& &  \multicolumn{2}{c|}{Supervision} & \multicolumn{4}{c|}{Sem ($mIoU$)} & Pan ($PQ$)\\
			\hline
			Method & Backbone & VL Pre-training & Training & A-847  & A-150 & P-459 & P-59 & ADE20K \\
			\hline
			\midrule			
			GroupViT \cite{xu2022groupvit} & ViT-S & - & 27M (caption) & - & - & - & 20.4 & \scalebox{0.85}[1]{$\times$}\\
			OpenSeg \cite{ghiasi2021open} & EfficientNet-B7 & 1.3B (ALIGN) & 240K (COCO Cap, COCO) &6.3 & 21.1 &9.0 &42.1 &\scalebox{0.85}[1]{$\times$} \\
			Xu \emph{et al.} \cite{xu2022simple} & Res-101 & 400M (CLIP) & 120K (COCO-stuff) & 7.0 & 20.5 & - & 47.7 & \scalebox{0.85}[1]{$\times$} \\
			MaskCLIP-panoptic \cite{ding2022open} & Res-50 & 400M (CLIP) & 120K (COCO Pan) &8.2 & 23.7 &10.0 &45.9 &15.1 \\
			OVSeg \cite{liang2022open} & Swin-B & 400M (CLIP) & 240K (COCO Cap, COCO-stuff) & 9.0 & 29.6 & 12.4 & 55.7 &\scalebox{0.85}[1]{$\times$} \\
			X-Decoder \cite{zou2022generalized} & Focal-T & - & 4M (caption, grounding, COCO Pan) & 9.2 & 29.6 & 16.1 & \textbf{64.0} & 21.8 \\
			ODISE \cite{xu2023open} & Swin-L & 400M (SD) + 400M (CLIP) & 240K (COCO Cap, COCO Pan) & 11.1 & 29.9 & 14.5 & 57.3 &22.6 \\
			\hline 
			Model A (modified Mask2Former \cite{cheng2022masked}) & Res-50 & - & 300K (grounding, LVIS base) & 7.6 & 22.0 & 9.8 & 42.2 & 16.8 \\
			UOVN (Ours) & Res-50 & - & 300K (grounding, LVIS base)  & \textbf{13.5} & \textbf{30.7} & \textbf{17.1} & 54.3 & \textbf{23.9} \\
			\bottomrule	
	\end{tabular}}
	\label{tab_result_ade}
\end{table*}	

\begin{table*}
	\caption{Zero-shot results on COCO. `Det' and `Ins' are detection and instance segmentation. We use the generalized zero-shot setting \cite{bansal2018zero}. `\scalebox{0.85}[1]{$\times$}' means the method does not deal with the task. Note that several methods \cite{feng2022promptdet, lin2022learning, rasheed2022bridging} use novel-class information during training, which is not practical for the OV task.}
	\centering
	\scalebox{0.93}{
		\begin{tabular}{l|c|c|c|ccc|ccc} 
			\toprule             
			& & & & \multicolumn{3}{c|}{Det ($AP^{50}$)} & \multicolumn{3}{c}{Ins ($AP^{50}$)} \\
			\hline
			Method & Backbone & VL Pre-training & Training & novel & base & all & novel & base & all \\
			\hline
			\midrule
			\emph{without novel classes:} & \ & \ & \ & \  & \ & \ & \ & \ & \ \\
			OVR-CNN \cite{zareian2021open} & Res-50 & - & 240K (COCO Cap, COCO base) &22.8 & 46.0 & 39.9 & \multicolumn{3}{c}{\scalebox{0.85}[1]{$\times$}} \\
			Detic \cite{zhou2022detecting} & Res-50 & 400M (CLIP) & 240K (COCO Cap, COCO base) &27.8 & 47.1 & 45.0 & \multicolumn{3}{c}{\scalebox{0.85}[1]{$\times$}} \\
			ViLD \cite{gu2021open}         & Res-50 & 400M (CLIP) & 120K (COCO base) &27.6 & 59.5 & 51.3 & \multicolumn{3}{c}{\scalebox{0.85}[1]{$\times$}} \\
			XPM \cite{huynh2022open}      & Res-50 & - & 5.1M (caption, OI, COCO base) & 29.9 & 46.3 & 42.0 & 21.9 & 41.5 & 36.3 \\
			OV-DETR \cite{zang2022open}   & Res-50 & 400M (CLIP) & 120K (COCO base) & 29.4 & 61.0 & 52.7 & \multicolumn{3}{c}{\scalebox{0.85}[1]{$\times$}} \\
			RegionCLIP \cite{zhong2022regionclip} & Res-50 & 400M (CLIP) & 3M (CC, COCO base) & 31.4 & 57.1 & 50.4 & \multicolumn{3}{c}{\scalebox{0.85}[1]{$\times$}} \\
			F-VLM \cite{kuo2022f} & Res-50 & 400M (CLIP) & 120K (COCO base) & 28.0 & - & 39.6 & \multicolumn{3}{c}{\scalebox{0.85}[1]{$\times$}} \\
			GLIP \cite{li2022grounded} (re-train) & Res-50 & - & 320K (grounding, COCO base) & 27.4 & 56.2 & 48.7 & \multicolumn{3}{c}{\scalebox{0.85}[1]{$\times$}} \\
			Model A (modified Mask2Former \cite{cheng2022masked}) & Res-50 & - & 320K (grounding, COCO base) & \multicolumn{3}{c}{\scalebox{0.85}[1]{$\times$}} & 24.5 & 51.7 & 44.6 \\
			UOVN (Ours) & Res-50 & - & 320K (grounding, COCO base) & \textbf{32.7} & \textbf{62.6} & \textbf{54.2} & \textbf{31.1} & \textbf{54.8} & \textbf{48.6} \\
			\hline
			\emph{with novel classes:} & \ & \ & \ & \  & \ & \ & \ & \ & \ \\
			PromptDet \cite{feng2022promptdet} & Res-50 & 400M (CLIP) & 400M (LAION, COCO) & 26.6 & - & 50.6  & \multicolumn{3}{c}{\scalebox{0.85}[1]{$\times$}} \\
			VLDet \cite{lin2022learning} & Res-50 & 400M (CLIP) & 240K (COCO Cap, COCO) & 32.0 & 50.6 & 45.8 & \multicolumn{3}{c}{\scalebox{0.85}[1]{$\times$}} \\
			Rasheed \emph{et al.} \cite{rasheed2022bridging} & Res-50 & 400M (CLIP, MViT) & 240K (COCO Cap, COCO) & 36.6 & 54.0 & 49.4 & \multicolumn{3}{c}{\scalebox{0.85}[1]{$\times$}} \\
			\bottomrule	
	\end{tabular}}
	\label{tab_result_coco_zs}
\end{table*}

\subsection{Main Results}
We first report the cross-dataset results on COCO in Table~\ref{tab_result_coco}. Since prior works \cite{gu2021open, huynh2022open, ding2022open} usually use Res-50 as their backbones, we also use the Res-50 backbone for a fair comparison. GLIP \cite{li2022grounded} is the first work that reformulates object detection to referring grounding. However, it is trained with 27 million data as a vision-language pre-training. Rather than designing a pre-training, our work aims to provide an architecture for unified OV dense prediction. Therefore, we reproduce GLIP \cite{li2022grounded} with our 300K training data. 
Compared with GLIP \cite{li2022grounded}, our method obtains 5.9\% gains on the detection task. Moreover, GLIP \cite{li2022grounded} is only designed for detection, while our method is generic and can generate various segmentation results. We design another baseline called `Model A' where we add a text encoder to Mask2former \cite{cheng2022masked} and replace the classification part in Mask2former \cite{cheng2022masked} with a vision-language similarity matrix for OV recognition. This model can only deal with segmentation tasks and is trained on segmentation data. Due to referring segmentation data being relatively less, the performance of `Model A' is impacted. Our unified network flexibly combines referring detection and segmentation data for training, and we also propose MMM decoding and UOVN training mechanisms. Therefore, our method achieves significant improvements as well as outperforms most of other state-of-the-art approaches. The original GLIP \cite{li2022grounded} performs better because of a larger backbone and millions of training/pre-training data. X-Decoder \cite{zou2022generalized} and ODISE \cite{xu2023open} report superior COCO segmentation results, but they are fully-supervised for COCO.

Table \ref{tab_result_ade} shows the segmentation results on ADE20K and Pascal Context. GLIP \cite{li2022grounded} cannot generate segmentation results. 
Our method brings remarkable improvements, compared to baseline `Model A'. X-Decoder \cite{zou2022generalized} and ODISE \cite{xu2023open} show the best performance in prior works, while our method outperforms them under most metrics, especially for a mass of classes (e.g., 847 and 459 classes). These results demonstrate our OV ability, bulit upon our proposed MMM decoding to integrate multi-modal OV data and adaptation losses to reduce cross-dataset gaps.
Instead of OV recognition, X-Decoder \cite{zou2022generalized} focuses on segmentation, and uses more segmentation labels as well as an elaborate backbone. Therefore, it shows higher IoU when segmenting 59 classes, which are almost covered by their segmentation labels.

\begin{figure*} [t]
	\centerline{\includegraphics[scale=0.39]{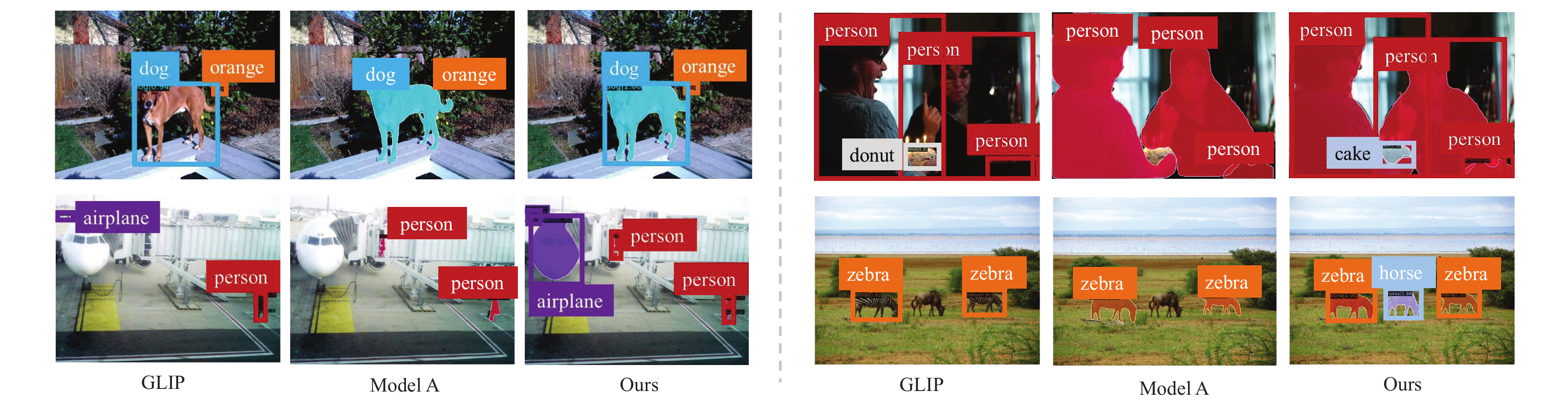}}
	\caption{Qualitative cross-dataset object detection and instance segmentation results on COCO validation. Re-trained GLIP \cite{li2022grounded} and `Model A' ignore and mis-classify objects, such as the `cake' in the upper right image as well as the `horse' in the bottom right image. Our method can detect, segment and recognize these objects.}
	\label{fig_results1}
\end{figure*}

\begin{table*}
	\caption{Zero-shot results on LVIS. `Det' and `Ins' are detection and instance segmentation. `\scalebox{0.85}[1]{$\times$}' means this method does not tackle this task. Note that several methods \cite{feng2022promptdet, lin2022learning, rasheed2022bridging} use novel-class information during training, which is not practical for the OV task.}
	\centering
	\scalebox{0.8}{
		\begin{tabular}{l|c|c|c|cccc|cccc} 
			\toprule             
			& & & & \multicolumn{4}{c|}{Det ($mAP$)} & \multicolumn{4}{c}{Ins ($mAP$)} \\
			\hline
			Method & Backbone & VL Pre-training & Training & rare & common & frequent & all & rare & common & frequent & all \\
			\hline
			\midrule
			\emph{without novel classes:} & \ & \ & \ & \  & \ & \ & \ & \ & \ & \ & \  \\
			Detic \cite{zhou2022detecting} & Res-50 & 400M (CLIP) & 240K (COCO Cap, COCO base) & \multicolumn{4}{c|}{\scalebox{0.85}[1]{$\times$}} & 17.8 & 26.3 & 31.6 & 26.8 \\
			ViLD \cite{gu2021open}         & Res-50 & 400M (CLIP) & 100K (LVIS base) & \multicolumn{4}{c|}{\scalebox{0.85}[1]{$\times$}} & 16.6 & 24.6 & 30.3 & 25.5 \\
			DetPro \cite{du2022learning}  & Res-50 & 400M (CLIP) & 100K (LVIS base) & 20.8 & 27.8 & 32.4 & 28.4 & 19.8 & 25.6 & 28.9 & 25.9  \\
			OV-DETR \cite{zang2022open}   & Res-50 & 400M (CLIP) & 100K (LVIS base) & \multicolumn{4}{c|}{\scalebox{0.85}[1]{$\times$}} & 17.4 & 25.0 & 32.5 & 26.6 \\
			RegionCLIP \cite{zhong2022regionclip} & Res-50 & 400M (CLIP) & 3M (CC,LVIS base) & \multicolumn{4}{c|}{\scalebox{0.85}[1]{$\times$}} & 17.1 & 27.4 & 34.0 & 28.2 \\
			F-VLM \cite{kuo2022f} & Res-50 & 400M (CLIP) & 100K (LVIS base) & \multicolumn{4}{c|}{\scalebox{0.85}[1]{$\times$}} & 18.6 & - & - & 24.2\\
			GLIP \cite{li2022grounded} (re-train) & Res-50 & - & 300K (grounding, LVIS base) & 18.9 & 27.0 & 31.2 & 26.9 & \multicolumn{4}{c}{\scalebox{0.85}[1]{$\times$}} \\
			Model A (modified Mask2Former \cite{cheng2022masked}) & Res-50 & - & 300K (grounding, LVIS base) & \multicolumn{4}{c|}{\scalebox{0.85}[1]{$\times$}}  & 15.7 & 28.1 & 33.4 & 27.5 \\
			UOVN (Ours) & Res-50 & - & 300K (grounding, LVIS base) & \textbf{22.6} & \textbf{31.9} & \textbf{37.1} & \textbf{32.0} & \textbf{22.0} & \textbf{30.7} & \textbf{35.7} & \textbf{30.9} \\
			\hline
			\emph{with novel classes:} & \ & \ & \ & \  & \ & \ & \ & \ & \ & \ & \  \\
			PromptDet \cite{feng2022promptdet} & Res-50 & 400M (CLIP) & 400M (LAION, LVIS) & \multicolumn{4}{c|}{\scalebox{0.85}[1]{$\times$}} & 21.4 & 23.3 & 29.3 & 25.3\\
			VLDet \cite{lin2022learning} & Res-50 & 400M (CLIP) & 3M (CC, LVIS) & \multicolumn{4}{c|}{\scalebox{0.85}[1]{$\times$}} & 21.7 & 29.8 & 34.3 & 30.1\\
			Rasheed \emph{et al.} \cite{rasheed2022bridging} & Res-50 & 400M (CLIP, MViT) & 1.5M (ImageNet21K, LVIS) & \multicolumn{4}{c|}{\scalebox{0.85}[1]{$\times$}} & 21.1 & 25.0 & 29.1 & 25.9\\
			\bottomrule	
	\end{tabular}}
	\label{tab_result_lvis_zs}
\end{table*}

Table \ref{tab_result_coco_zs} and Table \ref{tab_result_lvis_zs} report the zero-shot results on COCO and LVIS. UOVN achieves significant improvements, compared to other state-of-the-art methods with the same backbone.
Note that several works \cite{feng2022promptdet, lin2022learning, rasheed2022bridging} use a different setting that involves novel-class information of the target dataset. Concretely, they generate pseudo training samples based on novel classes of the target dataset, which is not practical from the OV perspective. Nevertheless, our method also outperforms most of them. All these superior results demonstrate the effectiveness of our unified network, as well as our proposed MMM decoding and adaptation strategies.

\subsection{Discussion} 
In this section, we conduct a series of discussions to further verify the effectiveness of every module in our UOVN, including (1) the effects of each main component, (2) different settings of the MMDA module and (3) UOVN training, (4) task analysis, (5) the effects of prompt, as well as (6) the analysis of qualitative results.

\textbf{Main components.} We report ablation study results in Table \ref{tab_ablation}. Compared with our baseline `Model A', our MMM decoding first incorporates a detection branch (`Model C') which allows joint training on both detection and segmentation data. Through the joint training, the instance and panoptic segmentation results are increased by 5.4\% and 3.4\%. The MMDA module (`Model D') in our decoding yields improvements of 2.6\%, 2.2\% and 3.0\% on the detection, instance and panoptic segmentation tasks, respectively. The UOVN training strategy (`Full Model') improves the performance by 1.7\%, 2.1\% and 1.9\% for detection, instance and panoptic segmentations. These results demonstrate the effectiveness of our MMM decoding, MMDA module and UOVN training.

\begin{table*}
	\caption{Ablation study on COCO validation at the cross-dataset scenario. `Det', `Ins' and `Pan' mean detection, instance and panoptic segmentations. `\scalebox{0.85}[1]{$\times$}' indicates that the method cannot tackle this task. In `Model A', we use a vision-language similarity matrix to replace classification layers in Mask2Former \cite{cheng2022masked} for OV recognition.}
	\centering
	\scalebox{1.0}{
		\begin{tabular}{l|cccc|c|c|ccc}
			\toprule
			& Segmentation & Detection & MMDA & UOVN        & Det  & Ins & Pan\\
			Method & tasks        & task      & module & training & $mAP$ & $mAP$ & $PQ$ \\
			\midrule
			Model A (modified Mask2Former \cite{cheng2022masked}) & \checkmark & & & & \scalebox{0.85}[1]{$\times$} & 20.8 & 24.4 \\
			Model B & & \checkmark & & & 33.8  & \scalebox{0.85}[1]{$\times$} & \scalebox{0.85}[1]{$\times$} \\
			Model C & \checkmark & \checkmark & & & 37.0 & 26.2 & 27.8 \\
			Model D & \checkmark & \checkmark & \checkmark & & 39.6 & 28.4 & 30.8 \\
			Full Model  & \checkmark & \checkmark & \checkmark & \checkmark & \textbf{41.3}  & \textbf{30.5} & \textbf{32.7} \\    
			\bottomrule	
	\end{tabular}}
	\label{tab_ablation}
\end{table*}

\begin{table}
	\caption{The effects of different settings in our MMDA module. All results are on COCO validation at the cross-dataset scenario. `Det', `Ins' and `Pan' mean detection, instance and panoptic segmentations.}
	\centering
	\scalebox{0.9}{
		\begin{tabular}{l|cc|c|c|ccc}
			\toprule
			& Language-specific & Joint vision-    & Det  & Ins & Pan\\
			Method &  offsets        & language attention  & $mAP$ & $mAP$ & $PQ$\\
			\midrule
			Model C &  &  & 37.0 & 26.2 & 27.8  \\
			Model E & \checkmark &  & 38.1 & 26.7 & 29.2 \\
			Model F &  & \checkmark & 38.8 & 27.6 & 30.3  \\
			Model D & \checkmark & \checkmark & \textbf{39.6} & \textbf{28.4} & \textbf{30.8}  \\
			\bottomrule	
	\end{tabular}}
	\label{tab_mmda}
\end{table}

\begin{table}
	\caption{The effects of different settings in our UOVN training. All results are on COCO validation at the cross-dataset scenario. `Det', `Ins' and `Pan' mean detection, instance and panoptic segmentations.}
	\centering
	\scalebox{0.9}{
		\begin{tabular}{l|cc|c|c|ccc}
			\toprule
			&  Global adaptation &  Local adaptation  & Det  & Ins & Pan\\
			Method &  $L_{adp-g}$  & $L_{adp-l}$ & $mAP$ & $mAP$ & $PQ$ \\
			\midrule
			Model D &  &  & 39.6 & 28.4 & 30.8  \\
			Model G & \checkmark &  & 40.4 & 28.9 & 31.5 \\
			Model H &  & \checkmark & 40.7 & 30.2 & 32.1  \\
			Full Model  & \checkmark & \checkmark & \textbf{41.3}  & \textbf{30.5} & \textbf{32.7}  \\
			\bottomrule	
	\end{tabular}}
	\label{tab_uovn_training}
\end{table}

\textbf{MMDA module.} We then dissect our MMDA module. The results are shown in Table \ref{tab_mmda}. In `Model E', we only use language query to guide the deformable offest generation. In `Model F', we exclude language features from the selected feature set and only use vision cues to calculate attentions. Compared with the model without MMDA (`Model C'), both `Model E' and `Model F' increase the performance of all tasks, because our language-specific offsets and joint vision-language attention in MMDA better learn visual embedding from language cues. The entire MMDA module (`Model D') further improves the performance.

\textbf{UOVN training.} As our training data comes from diverse domains and tasks, we propose a global adaptation loss $L_{adp-g}$ and a local adaptation loss $L_{adp-l}$ to reduce domain gaps. Table \ref{tab_uovn_training} shows the effects of $L_{adp-g}$ and $L_{adp-l}$. Compared with `Model D', our $L_{adp-g}$ and $L_{adp-l}$ improve the detection performance by 0.8\% and 1.1\%, respectively. These losses also show improvements on other tasks. Meanwhile, the local adaptation loss $L_{adp-l}$ yields more improvements than the global adaptation loss $L_{adp-g}$, because local object embeddings are more crucial for dense prediction tasks. By combining global and local losses, our final model obtains the best performance.

\textbf{Task analysis.} In table \ref{tab_ablation}, `Model A' and `Model B' separately use segmentation and detection data for training. Compared with these task-specific models, the unified network `Model C' yields significant improvements, especially for segmentation tasks, because referring segmentation training data is relatively less than the data for referring detection.

\textbf{Prompt.} Language prompt is a frequently-used technology to enhance language representations in open-vocabulary methods. As described in Section \ref{Ecoding}, we use the CLIP prompt `A photo of [query]' during inference. Table \ref{tab_prompt} shows the results of our method without this prompt. It is seen that prompt slightly improves our performance for all tasks. We also use this prompt in other ablation models (i.e., `Model A-H' and `Full Model').

\begin{figure} [t]
	\centerline{\includegraphics[scale=0.30]{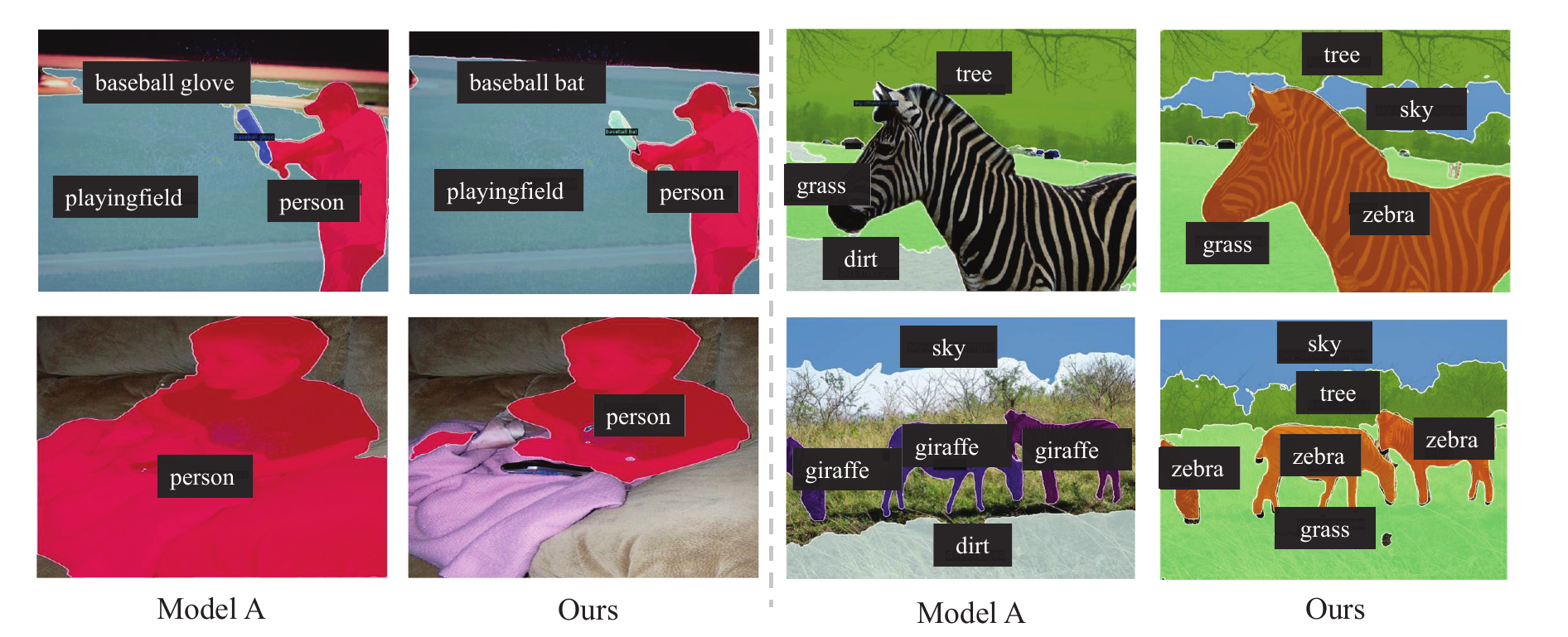}}
	\caption{Qualitative cross-dataset panoptic segmentation results on COCO validation. `Model A' does not well segment some objects, such as the `person' in the bottom left image and the `zebra', `tree' as well as `grass'in the bottom right image. Meanwhile, the `baseball bat' in the upper left image is mis-classified as `baseball glove'. Our method avoids these errors.}
	\label{fig_results2}
\end{figure}

\begin{table} [t]
	\caption{The effects of prompt on COCO validation. `Det', `Ins' and `Pan' mean object detection, instance and panoptic segmentations.}
	\centering
	\scalebox{1.0}{
		\begin{tabular}{l|C{1.5cm}|C{1.5cm}|ccc}
			\toprule
			& Det  & Ins & Pan\\
			Method & $mAP$ & $mAP$ & $PQ$ \\
			\midrule
			Ours without prompt & 40.5  & 29.6 & 31.9 \\
			Ours with prompt & \textbf{41.3}  & \textbf{30.5} & \textbf{32.7} \\
			\bottomrule	
	\end{tabular}}
	\label{tab_prompt}
\end{table}

\textbf{Qualitative results.} We depict qualitative results in Fig.~\ref{fig_results1} and Fig.~\ref{fig_results2}. In Fig.~\ref{fig_results1}, both re-trained GLIP \cite{li2022grounded} and `Model A' ignore the largest `airplane' in the bottom left image and the `horse' in the bottom right image. Moreover, GLIP \cite{li2022grounded} also mis-classifies `cake' as `donut' in the upper right image, and `Model A' cannot segment this object. Different from them, our method successfully detects, segments and recognizes these objects. A reason is that we combine segmentation and detection to recognize objects, and our MMM decoding, MMDA module as well as UOVN training better integrate multi-task, multi-modal and multi-dataset information. Fig.~\ref{fig_results2} shows the panoptic segmentation results. GLIP \cite{li2022grounded} is unable to tackle this task. `Model A' cannot segment the `zebra' as well as `sky' in the upper right image, and mis-recognizes `grass' as `dirt'. Our method reduces these mistakes, because we incorporate detection data for training, use MMM decoding for better OV object recognition, and propose adaptation to enhance the training.

\section{Conclusion}
In this paper, we have presented UOVN, a unified network for open-vocabulary object detection, semantic, instance and panoptic segmentations. We first introduce a multi-modal, multi-scale and multi-task (MMM) decoding mechanism to recognize open-category objects and generate bounding boxes as well as diverse segmentation masks. In MMM decoding, we propose a multi-modal multi-scale deformable attention (MMDA) module to leverage multi-modal information to enhance extracted embeddings. Secondly, a UOVN training mechanism, including global and local adaptation losses, is presented to reduce domain and task gaps among heterogeneous training data. Extensive experimental results on the COCO, LVIS, ADE20K and Pascal Context datasets demonstrate the effectiveness of our method.

\bibliographystyle{IEEEtran}
\bibliography{my_reference}
\end{document}